\documentclass{article}

\usepackage{arxiv}

\usepackage[utf8]{inputenc} 
\usepackage[T1]{fontenc}    
\usepackage{hyperref}       
\usepackage{url}            
\usepackage{booktabs}       
\usepackage{amsfonts}       
\usepackage{nicefrac}       
\usepackage{microtype}      
\usepackage{lipsum}
\usepackage{graphicx}
\usepackage{bm}
\usepackage{mathtools}
\usepackage{algorithm}
\usepackage{algorithmicx}
\usepackage{algpseudocode}
\usepackage{amsmath,amsthm,amssymb}
\DeclarePairedDelimiterX{\infdivx}[2]{(}{)}{%
  #1\;\delimsize|\delimsize|\;#2%
}
\newcommand{\kld}[2]{\ensuremath{D_{KL}\infdivx{#1}{#2}}}

\DeclareDocumentCommand \expectation { o m } {%
  \ensuremath{\mathbb{E}%
  \IfValueTF {#1} {%
    _{#1} \left[ #2 \right]%
  }{%
    \left[ #2 \right]%
  }%
  }%
}

\title{Organization of a Latent Space structure in VAE/GAN trained by navigation data }

\author{
  Hiroki Kojima\\
  University of Tokyo\\
  \texttt{kojima@sacral.c.u-tokyo.ac.jp} \\
   \And
 Takashi Ikegami \\
  University of Tokyo\\
  \texttt{ikeg@sacral.c.u-tokyo.ac.jp} \\
}

\begin{document}
\maketitle

\begin{abstract}
We present a novel artificial cognitive mapping system using generative deep neural networks, called variational autoencoder/generative adversarial network (VAE/GAN), which can map input images to latent vectors and generate temporal sequences internally. The results show that the distance of the predicted image is reflected in the distance of the corresponding latent vector after training. This indicates that the latent space is self-organized to reflect the proximity structure of the dataset and may provide a mechanism through which many aspects of cognition are spatially represented. The present study allows the network to internally generate temporal sequences that are analogous to the hippocampal replay/pre-play ability, where VAE produces only near-accurate replays of past experiences, but by introducing GANs, the generated sequences are coupled with instability and novelty.

\end{abstract}

\keywords{Cognitive Map \and GAN \and Place Cell \and Prediction \and Latent Space \and Chaos}

The cognitive map was first initially proposed by Tolman to explain the deliberate behavior of rats in a maze \cite{tolman1948cognitive}, and  place cells \cite{o1971hippocampus} and grid cells \cite{hafting2005microstructure} in the hippocampus are regarded as the biological implementation of it. The mechanism of the formation of these spatial representations has already been discussed \cite{moser2017spatial}, and today, the most prevalent explanations are based on the "path integration," which insists that the constructed spatial representation was based on its own movement and was calculated by integrating the self-motion information\cite{mcnaughton1996deciphering}. These views are supported by the findings that the place cells maintained stable in the dark environment\cite{quirk1990firing}, but lost the spatial selectivity by the suppression of the vestibular input \cite{stackman2002hippocampal} or in virtual reality (VR) environments \cite{aghajan2015impaired}.

However, some researchers questioned this path integration view, for example, claiming that grid cells are not necessary for the emergence of the place cells \cite{brandon2011reduction,koenig2011spatial}, implying that the spatial metric was constructed from the self-motion cues and was not the basis of the spatial representation. Also, the spatial representations varied among different species, for example, the spatial representation of monkeys is associated with egocentric spatial views, suggesting that the cognitive map was more related to the visually centered space in some species\cite{rolls2006spatial}. The manner in which cognitive maps are organized differ how Google Maps for example are organized. Maps are cognitive in the sense that they are correlated with the degree and complexity of our cognitive capability \cite{buzsaki2013memory}. Recently, many researches have reported that the hippocampal structure mapped not only the spatial structure but also the nonspatial features \cite{schiller2015memory,bellmund2018navigating,behrens2018cognitive}, such as the frequency of tones \cite{aronov2017mapping}, the bird cartoons \cite{constantinescu2016organizing}, and social relationships\cite{tavares2015map}. These findings suggest the existence of a more general underlying mapping mechanism, which is not confined to spatial processing based on the integration of self-motion signals. Furthermore, the hippocampus is not only the center of processing spatial representations, but also responsible for memories, especially episodic memories \cite{schiller2015memory}. In this context, the neural basis of episodic memory is known to be related to the imagery ability \cite{hassabis2007using,hassabis2007patients}, and it has been proposed that the many aspects of the memory functions of the hippocampus are considered to be related with scene reconstruction \cite{hassabis2007deconstructing}, highlighting the importance of the generative nature of the hippocampus.

Herein, as an alternative mechanism to construct the cognitive map, we hypothesize that the map is self-organized at the bottleneck of the sensory reconstruction system. To confirm this hypothesis, we implemented an artificial cognitive map system using only visual information and no explicit metric information. Our hypothesis was partly inspired in a study on generative deep neural networks, in which the generated images were smoothly mapped in the latent spaces, which are the bottleneck of the architecture \cite{kingma2013auto}.
Specifically, we used generative deep neural network called variational autoencoder / generative adversarial network (VAE/GAN)\cite{Larsen}, which is a combination of VAE \cite{kingma2013auto} and GAN \cite{Goodfellow,Radford}. These generative deep neural networks learns to encode input images into latent vectors and generate images from the vectors. The structure of the latent space is assumed to reflect the input data structure, for example, the "World model" simulation \cite{ha2018world} utilized this type of latent space vectors to encode visual inputs. We trained VAE/GAN on a first-person navigation task wherein the agent moves through a simple virtual environment and studies the characteristics of the internal representations. 

The approach of constructing an artificial cognitive map system to understand the underlying mechanism was first proposed by R{\"o}ssler \cite{rossler1981artificial}. The approach was subsequently realized by using recurrent neural networks (RNN). For example, Nolfi and Tani \cite{nolfi1999extracting} trained a robot with hierarchical RNNs such that the next sensory state of the robot can be predicted and found that each layer of the neural networks encode some regularities of the environment. Noguchi et al. \cite{noguchi2017cognitive} also trained a hierarchical RNNs with recurrent gated units using visual and motor information and found that the map of the environment was self-organized at the higher layer. Banino et al. \cite{banino2018vector} used a RNNs with a long short-term memory (LSTM) architecture and showed that the grid-cell-like structure was self-organized after the path integration tasks were trained.
These systems and other recent systems \cite{whittington2020tolman,uria2020spatial,rikhye2020learning,recanatesi2021predictive} have different structures, but all of them used motor commands as inputs and were basically based on path integration mechanisms. This is completely different from our system, which only uses visual inputs.

Our system was constructed to predict the upcoming frame of the input videos; thus, the systems can be regarded as an example of video prediction systems \cite{oprea2020,babaeizadeh2017stochastic,lee2018stochastic}. These previous studies basically focused on the quality of video predictions, but our research interest was the resulting structure in the latent space. Thus, we kept our system as simple as possible and quantitatively characterized the latent space vectors.

The other important aspect of the place cells in the hippocampus is "replay."  For instance, the cognitive map is activated not only when the mouse is actually exploring the environment but also when the mouse is replaying their past experiences or dreaming about exploring the environment \cite{o1978hippocampus,buzsaki1983cellular,dragoi2011preplay}. These sequences were found to not simply be an exact replay of the past experience \cite{foster2017replay}. For example, the place cells corresponding to the locations that have not yet been traversed are also activated \cite{gupta2010hippocampal,stella2019hippocampal}.
Our system can also generate temporal sequences by iteratively predicting the next frame in a "closed-loop." Because GAN is used to produce the cognitive map, we have realistic non-existing image scenes, which might result in generating a temporal sequence that is different from the exact replay of the past experience. Hence, we investigated the dynamics of the closed-loop generation for different mapping conditions and aimed to provide a possible mechanism to generate sequences that differ from the exact replay.

\section{Methods}

\subsection{Predictive VAE/GAN}

Our system is based on VAE/GAN \cite{Larsen}. VAE/GAN containts three deep convolutional neural networks: the generator (Gen), discriminator (Dis), and encoder (Enc). (Table.~\ref{tab:architecture}) To stabilize the GAN training, we used Wassestein GAN with gradient penalty (wGAN-gp)\cite{gulrajani2017improved} for representing GAN. 

Those components are very briefly summarized below:

VAE: This is a type of unsupervised learning that includes two convolutional neural networks: Enc and Gen. The Enc encodes the input images into low dimensional latent vectors, and the Gen reconstructs the images from the encoded latent vectors. VAE is characterized by the use of a probability distribution (normal distribution) in the latent space to encode the input image.

GAN: This is a type of unsupervised learning in which two networks compete with each other to learn. One network is the Gen, which generates data from random noise input, and the other network is the Dis, which judges whether the input image is the data generated by the Gen or the real data. Namely, the Gen tries to deceive the Dis, and the Dis tries not to be deceived by the Gen.

wGAN: This compensates for the shortcomings of the Jensen--Shanon divergence (e.g., the loss of gradient) and uses a loss function based on the Wasserstein distance (i.e., earth-moving distance) instead. Practically, wGAN stabilizes the GAN training.

Because we are interested in predicting the following scene based on the current scene, we train the network to predict an image that is $\tau$ steps ahead (i.e., $\bm{x}(t+\tau)$) of the current image ($\bm{x}(t)$), by imposing the network $\mbox{Gen}(\mbox{Enc}(\bm{x}(t))) = \bm{x}(t+\tau)$ instead of $\mbox{Gen}(\mbox{Enc}(\bm{x}(t))) = \bm{x}(t)$, where $\mbox{Enc}(\bm{x})$ and $\mbox{Gen}(\bm{z})$ are the outputs of the Enc and the Gen respectively.

The actual loss functions used for training in this experiment are presented below, starting with the loss function of the VAE:

\begin{equation}
  \mathcal{L}_{\mbox{VAE}} = \mathcal{L}_{\mbox{prior}} + \mathcal{L}_{\mbox{llike}}^{\mbox{pixel}}
    \label{eq:loss_all_vae}
\end{equation}

with
\begin{eqnarray}
  \mathcal{L}_{\mbox{prior}} & = & \kld{q(\bm{z}|\bm{x})}{p(\bm{z})}\\
  \mathcal{L}_{\mbox{llike}}^{\mbox{pixel}} & = & -\expectation[q(\bm{z}|\bm{x}(t))]{\log p( \bm{x}(t+\tau)|\bm{z})}
    \label{eq:loss_vae}
\end{eqnarray}

where $q(\bm{z}|\bm{x})$ is the Enc, $p(\bm{x}|\bm{z})$ is the Gen, $D_{KL}(q||p)$ stands for Kullback-Leibler divergence, and $p(\bm{z}) = \mathcal{N}(\bm{0},\bm{I})$. 

The loss function of VAE/GAN uses the loss function from the Dis in addition to the VAE loss function $\mathcal{L}_{\mbox{VAE}}$ (Eq.~\ref{eq:loss_all_vae}) and is represented as follows:

\begin{equation}
  \mathcal{L}_{\mbox{VAE/GAN}} = \mathcal{L}_{\mbox{prior}} + \mathcal{L}_{\mbox{llike}}^{\mbox{pixel}} + \alpha \mathcal{L}_{\mbox{GAN}},
    \label{eq:loss_all_pixelLoss}
\end{equation}

with 
\begin{eqnarray}
\mathcal{L}_{\mbox{GAN}} & = & \mbox{Dis}(\bm{\hat{x}}) - \mbox{Dis}(\bm{x}) + \lambda \expectation[p(\bm{\hat{x}})]{(\| \nabla_{\bm{\hat{x}}}\mbox{Dis}(\bm{\hat{x}}) \|-1)^2},
\end{eqnarray}

where $\lambda = 10, \bm{\hat{x}} \sim p(\bm{x}|\bm{z})$, and $\alpha$ is the weight parameter for the GAN loss. $\mbox{Dis}(\bm{x})$ represents the outputs of the Dis. We set $\alpha = 1$, if not stated otherwise. 

In the VAE/GAN used in its first instance in the literature \cite{Larsen}, the reconstruction error of VAE/GAN was not the pixel loss between the input image and the generated image, but was measured using the middle layer activation (Appendix.1).

In our experiments, we set the dimension of the latent space to $d_z = 5, 10, 20$ and the prediction time step to $\tau = 0, 5, 30$ and compared the results of VAE (Eq.~\ref{eq:loss_all_vae}) and VAE/GAN (Eq.~\ref{eq:loss_all_pixelLoss}) under each condition. We trained the network for each condition three times by changing the random seed. We trained all models with Adam\cite{kingma2014adam} with $\beta_1 = 0, \beta_2 = 0.9$, and the learning rate of 0.0002 as an optimizer. The batch size was set to 64. The Dis was updated five times during each iteration, following the training procedure in wGAN-gp \cite{gulrajani2017improved}. We used Chainer \cite{tokui2015chainer} for the actual implementation and the network structures were based on the implementation by pfnet-research\cite{chainer-gan-lib}.

\begin{table}
  \centering
  \begin{tabular}{lll}
    \toprule

    Enc     & Gen     & Dis \\
    \midrule
    $3\times3$ 64 conv., leaky ReLU & fully-connected, BNorm, leaky ReLU & $3\times3$ 64 conv., leaky ReLU    \\
    
    $4\times4$ 128 conv., leaky ReLU & $4\times4$ 512 deconv., BNorm, leaky ReLU & $4\times4$ 128 conv., leaky ReLU \\
    
    $3\times3$ 128 conv., leaky ReLU & $4\times4$ 256 deconv., BNorm, leaky ReLU & $3\times3$ 128 conv., leaky ReLU \\
    
    $4\times4$ 256 conv., leaky ReLU & $4\times4$ 128 deconv., BNorm, leaky ReLU & $4\times4$ 256 conv., leaky ReLU\\
    
    $3\times3$ 256 conv., leaky ReLU  & $3\times3$ 64 deconv., BNorm, leaky ReLU & $3\times3$ 256 conv., leaky ReLU  \\
    
    $4\times4$ 512 conv., leaky ReLU  &  & $4\times4$ 512 conv., leaky ReLU  \\
        
    $3\times3$ 512 conv., leaky ReLU  &  & $3\times3$ 512 conv., leaky ReLU  \\
    
    fully-connected  &  & fully-connected  \\

    \bottomrule
    \\
  \end{tabular}

   \caption{The architecture of the components of our networks, the encoder (Enc), the generator (Gen) and the discriminator (Dis).}
\label{tab:architecture}
\end{table}

\subsection{Dataset}

\begin{figure}
\begin{tabular}{cc}
  \begin{minipage}[t]{0.7\hsize}
    \centering
  \includegraphics[width=\linewidth]{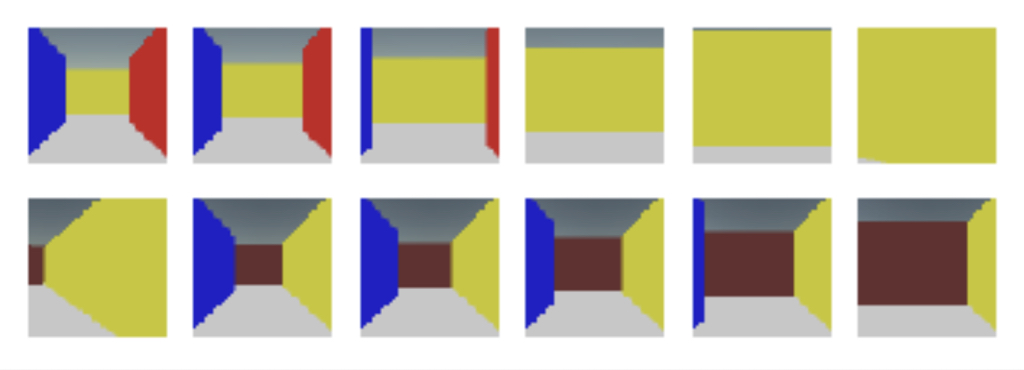}
  \end{minipage} &

 \begin{minipage}[t]{0.25\hsize}
   \centering
  \includegraphics[width=\linewidth]{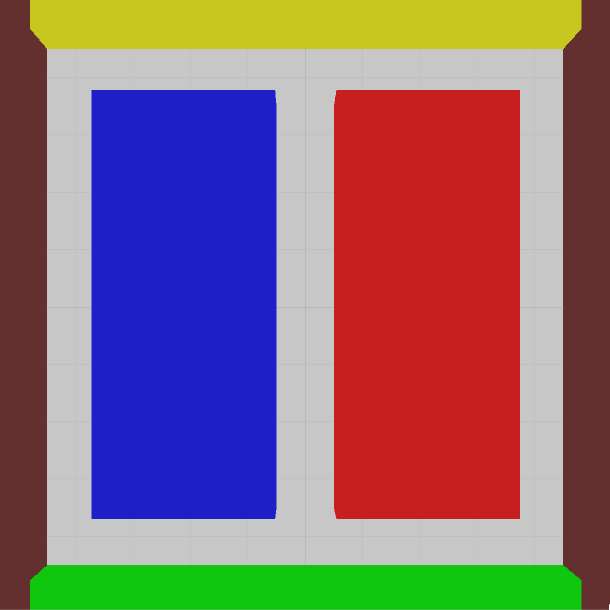}
  \end{minipage}

    \end{tabular}

  \caption{Examples of the temporal sequence of the images in our dataset. Right: Top view of the virtual environment. }
  \label{fig:fig1}
\end{figure}

As a training dataset, we used a series of first-person visual inputs of an agent moving in a virtual environment. For this purpose, we built a simple 3D virtual environment similar to a figure-8 maze using Unity and captured visual images of agents moving through it at a constant speed. The environment has one junction, at the intersection of the figure-8, and the agent randomly chooses which direction to go in. The captured images consisted of $64 \times 64$ pixels, for a total of 480 images (Fig.~\ref{fig:fig1}).

\section{Results}
\label{sec:result}

\begin{figure}
  \centering
  \includegraphics[width=\linewidth]{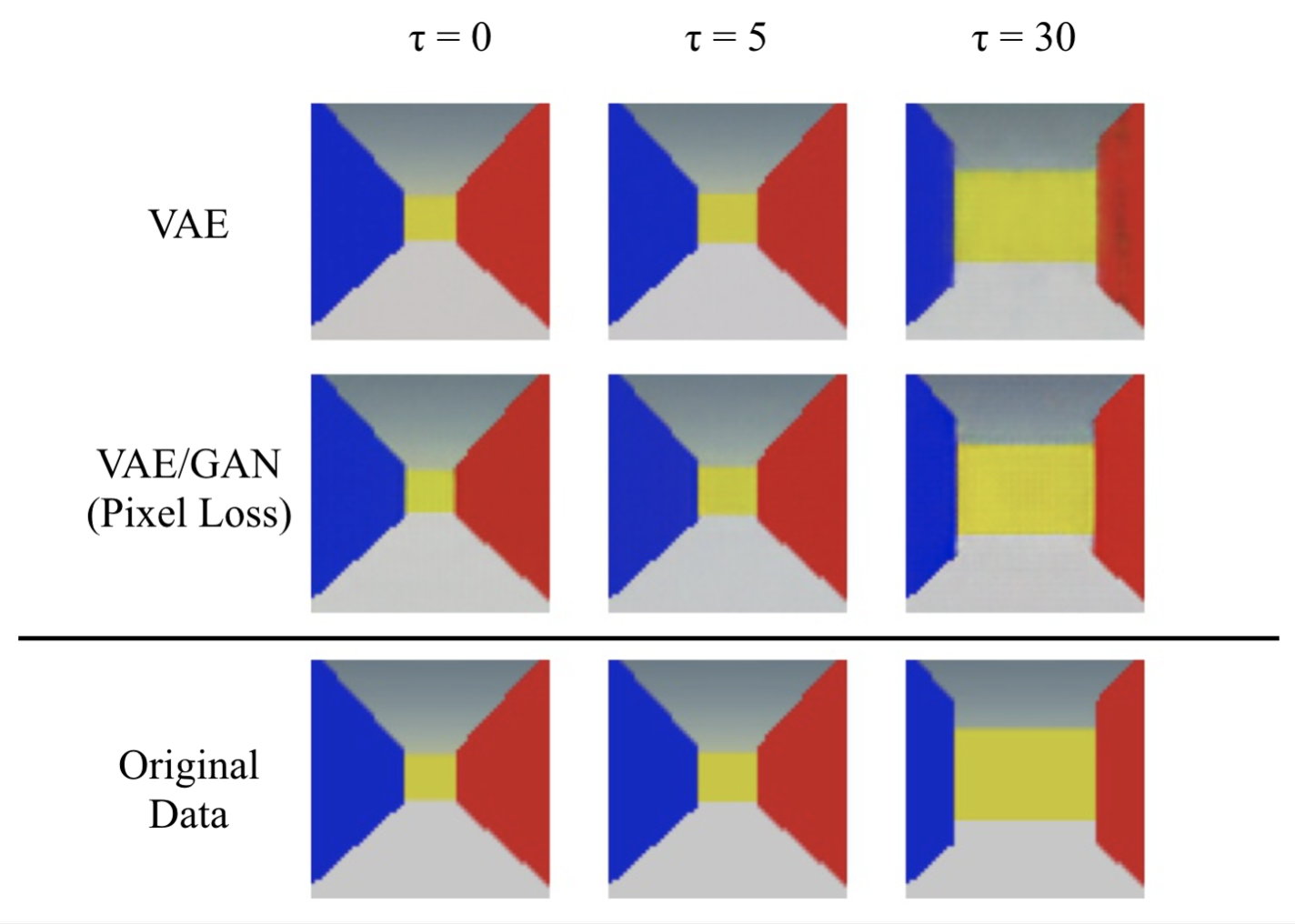}
  \caption{Examples of the generated images from the trained networks of VAE and VAE/GAN, with different prediction time steps ($\tau = 0,5,30$). The dimension of latent space was set to $z=10$. The targeted images are shown in the bottom row.} 
  \label{fig:fig2}
\end{figure}

Using these networks, we tested whether the agents could predict the next scene from the current one. The results of the training (Fig.~\ref{fig:fig2}) show that the network can successfully predict the scene $\tau$ steps in advance in each condition. The loss function was stabilized after around 8000 iterations of training in all conditions; thus, we will use the results after 8000 iterations in the following analysis. Unless otherwise stated, the analysis is based on the results averaged over 8000 to 10000 iterations from each seed (n = 3).
We have experimented with latent space dimensions $d_z = 5, 10, 20$, but in our analysis, the trend of the results did not change for varying values of $d_z$, so we will only describe the results for $d_z = 10$. We first analyzed how the images were mapped into the latent space by the Enc. Then, we introduced a "closed-loop image sequence generation" using the generated images as input recursively and characterized the sequences generated by the closed-loop method for each condition.

We first investigated the nature of the latent mappings. The following questions were addressed: Are images that are close in distance in a real space mapped into close points in the latent space (Fig.~\ref{fig:corr_distance})? How high is the dimensionality of latent space (Fig.~\ref{fig:pca})? Is the orbit properly captured (Fig.~\ref{fig:pca_mapping}, \ref{fig:pca_features})? Can a clear image be generated (Fig.~\ref{fig:PCA_grid})?
Then, by varying the ratio of VAE to GAN, we examine how the GAN contributes to the above properties, including the smoothness of the latent space and the similarities in the left and right pathways  (Fig.~\ref{fig:param}).
Furthermore, by adopting the predicted images of the network as input images in the next iteration, we can check how stable or unstable the representation of the latent space is as a deterministic dynamical system  (Fig.~\ref{fig:TrajType}).

\subsection{Patterns in Latent Spaces}

The Enc encodes the images $\bm{x}(t)$ in the dataset into the latent vectors as $\bm{z}(t) = \mathrm{Enc}(\bm{x}(t))$. We analyze the latent vector $\{\bm{z}(t)\}$ and investigate how the input images are mapped into the latent space.

\begin{figure}
  \centering
  \includegraphics[width=\linewidth]{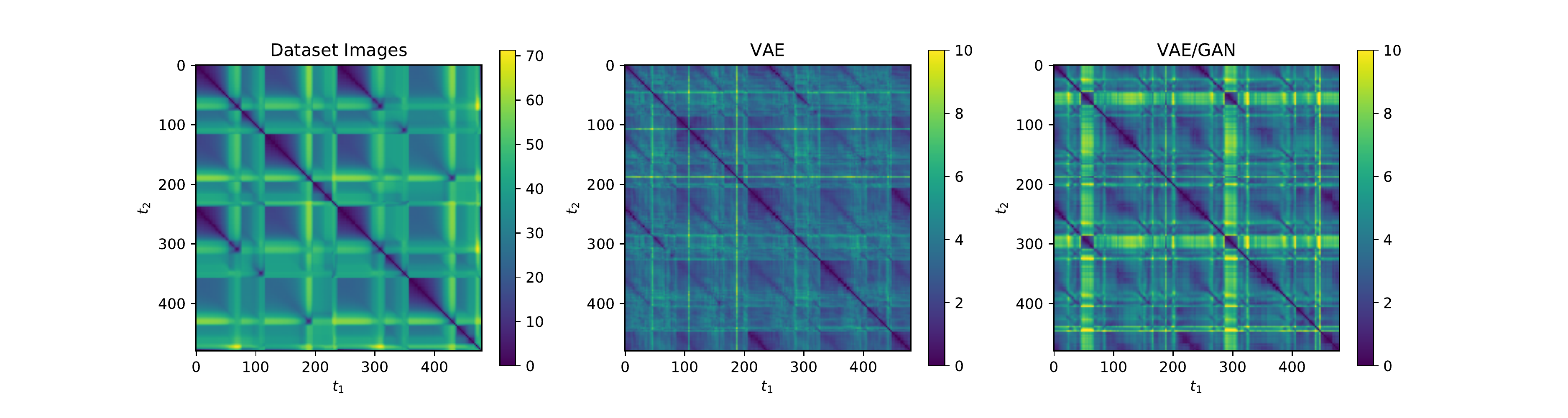}
  \caption{The distance matrix of image dataset $\| \bm{x}(t_1) - \bm{x}(t_2)\|$ (left) and the examples of the distance matrix of the corresponding latent vectors of VAE and VAE/GAN $\| \bm{z}(t_1) - \bm{z}(t_2)\|$ ($\tau = 30$) (middle and right). }
  \label{fig:DistanceExample}
\end{figure}

\begin{figure}
  \centering
  \includegraphics[width=\linewidth]{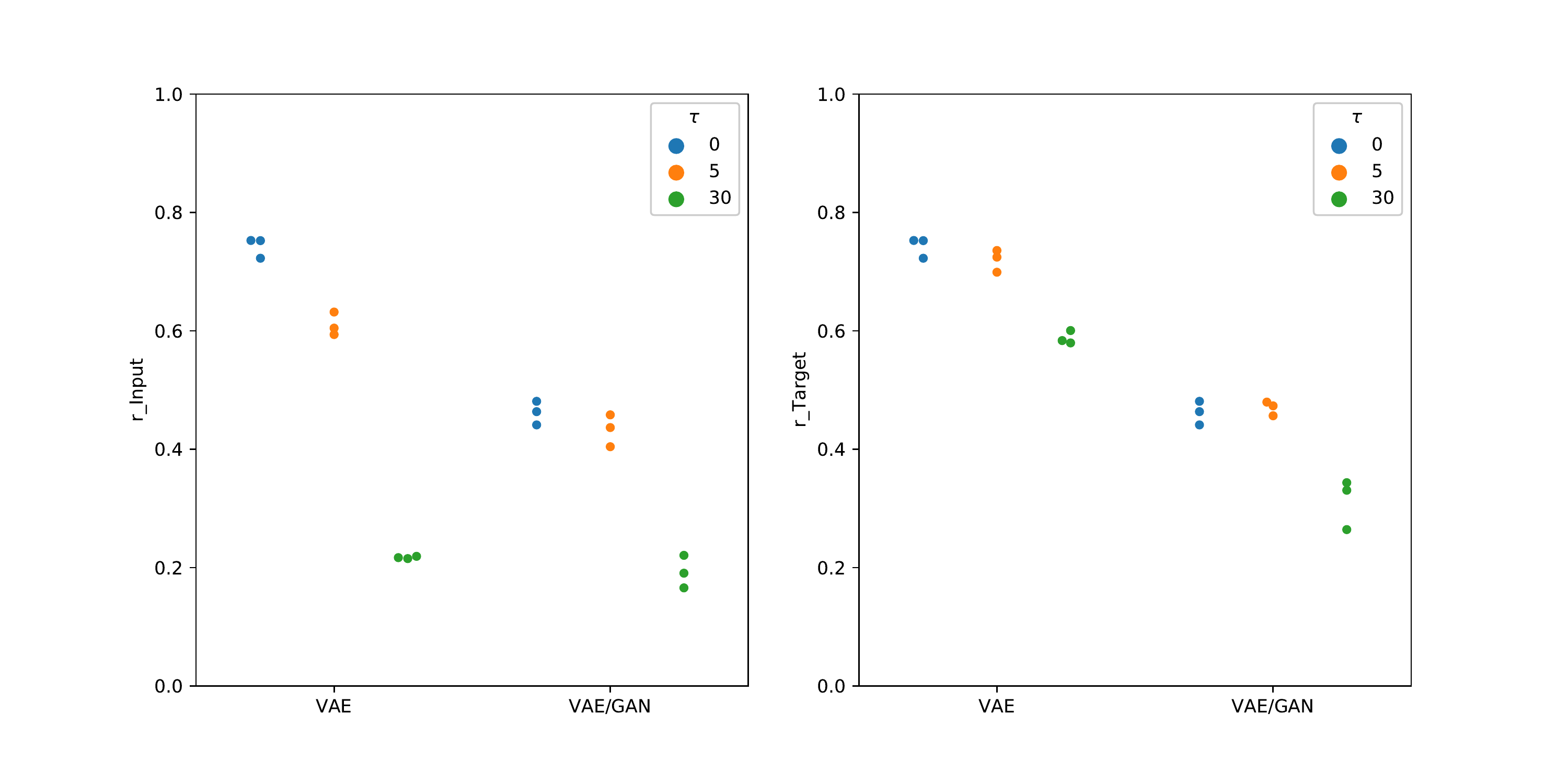}
  \caption{Correlation coefficient for distance structure of the  input images/ target images and the latent vectors. Each point correspond to the result from each random seed ($n=3$). Left: Correlation coefficient between the distance matrix of the input images $\| \bm{x}(t_1) - \bm{x}(t_2)\|$ and the distance matrix of the corresponding latent vectors $\| \bm{z}(t_1) - \bm{z}(t_2)\|$. Right: Correlation coefficient between the distance matrix of the target images $\| \bm{x}(t_1+\tau) - \bm{x}(t_2+\tau)\|$ and the distance matrix of the latent vectors $\| \bm{z}(t_1) - \bm{z}(t_2)\|$. }
  \label{fig:corr_distance}
\end{figure}

Since the agent traverses the maze while looking at the images, it is important to know the extent to which the actual image pattern is properly embedded in the latent space.
For this purpose, we compared the distance matrix of the input images $\| \bm{x}(t_1) - \bm{x}(t_2)\|$, the target images $\| \bm{x}(t_1 + \tau) - \bm{x}(t_2 + \tau)\|$ and the corresponding latent vectors $\| \bm{z}(t_1) - \bm{z}(t_2)\|$ and calculated the correlation coefficients between the values of these distance matrices (Fig.~\ref{fig:DistanceExample}). We found that a correlation exists and is stronger between the target images and the latent vectors than between the input images and the latent vectors, especially at $\tau = 30$ ($p < 0.001$, Tukey’s HSD) (Fig.~\ref{fig:corr_distance}).
This indicates that the distance structure of the latent vectors reflected the distance structure of the images after $\tau$ time steps.

\begin{figure}
  \centering
  \includegraphics[width=\linewidth]{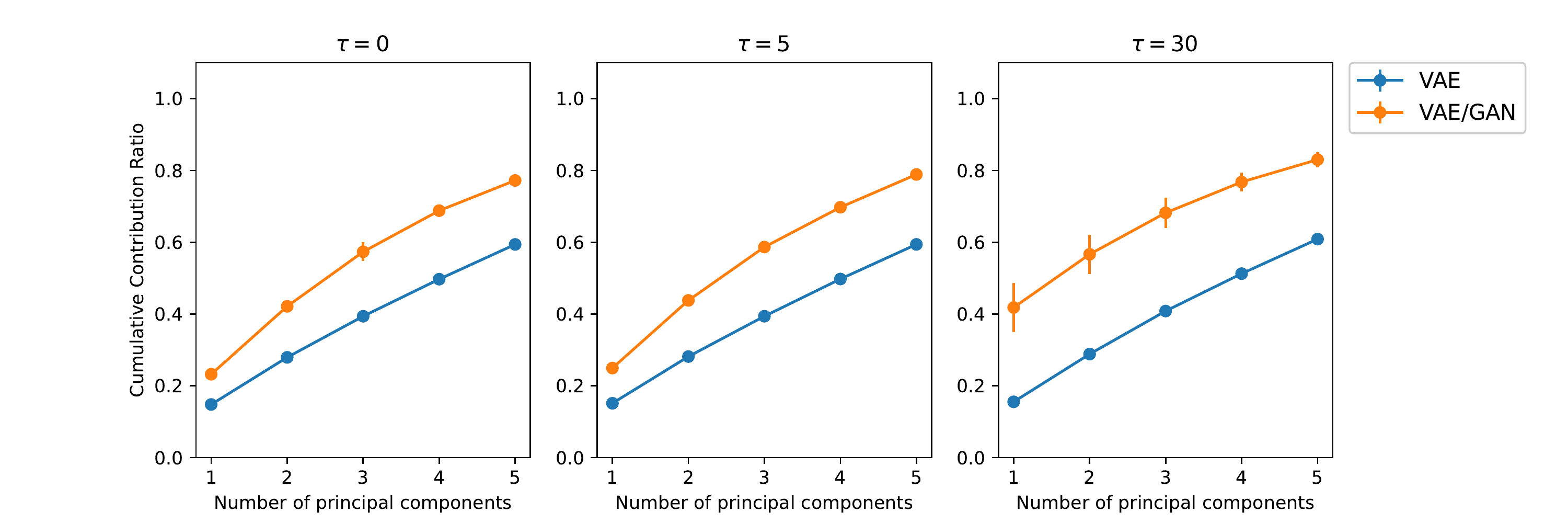}
  \caption{Cumulative contribution ratios of principal component analysis (PCA) ($d_z=10$) with different prediction timesteps, $\tau = 0, 5, 30$. The error bars denote the standard deviation of the results from different random seeds ($n=3$). }
  \label{fig:pca}
\end{figure}

\begin{figure}
\begin{tabular}{cc}
  \begin{minipage}[t]{0.7\hsize}
    \centering
  \includegraphics[width=\linewidth]{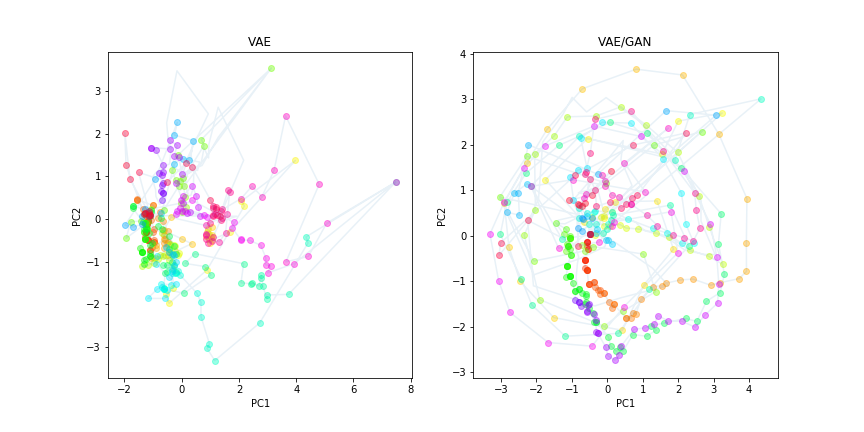}
  \end{minipage} &

 \begin{minipage}[t]{0.2\hsize}
   \centering
  \includegraphics[width=\linewidth]{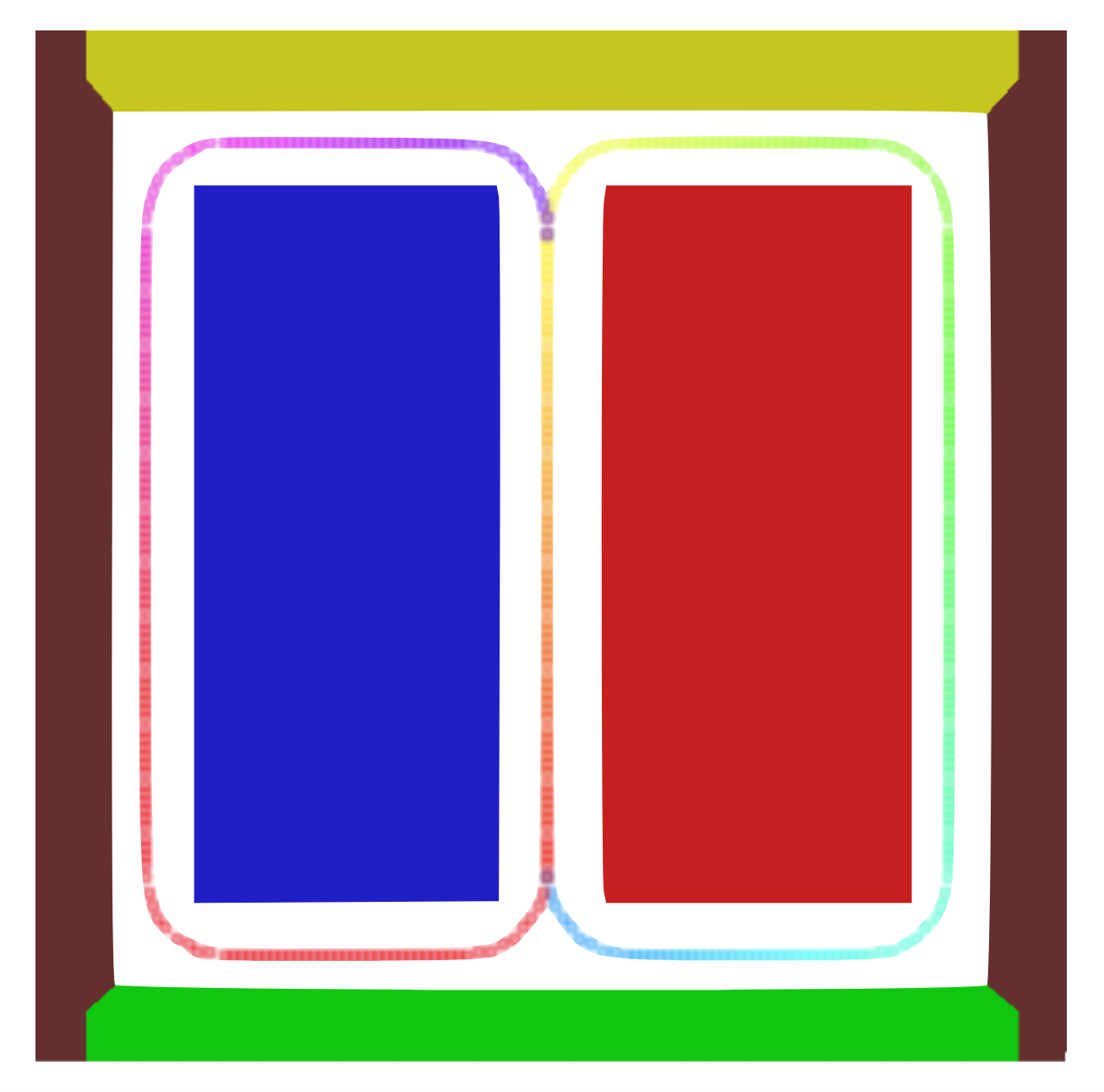}
  \end{minipage}

    \end{tabular}

  \caption{Example of the latent vectors $\{\bm{z}(t)\}$ mapped in the PCA space. Each color corresponds to the position in the virtual space (Right). Left: VAE ($\tau=5$), Middle: VAE/GAN ($\tau=5$). }
  \label{fig:pca_mapping}
\end{figure}

\begin{figure}
\begin{tabular}{cc}
  \begin{minipage}[t]{0.5\hsize}
    \centering
  \includegraphics[width=\linewidth]{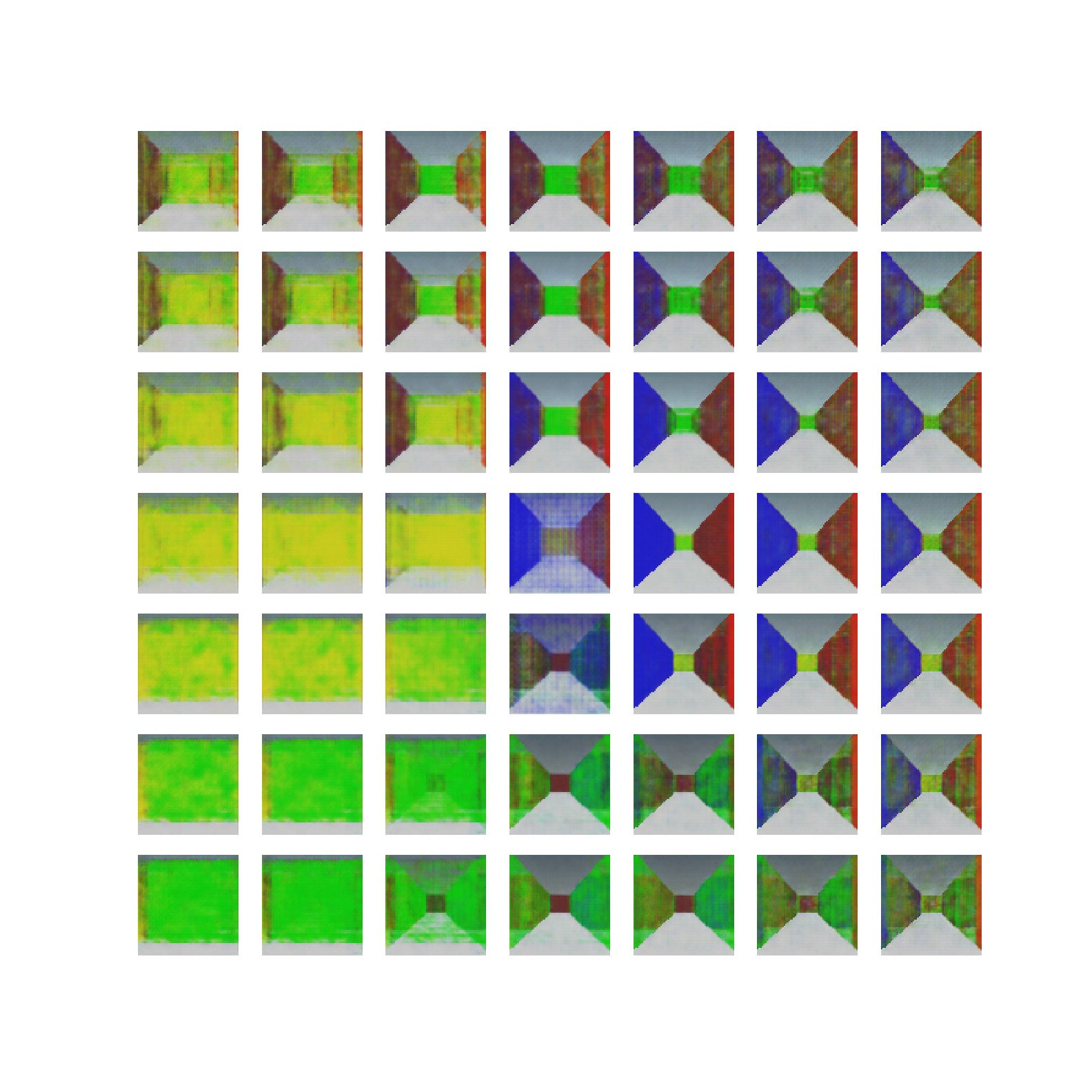}
  \end{minipage} &

 \begin{minipage}[t]{0.5\hsize}
   \centering
  \includegraphics[width=\linewidth]{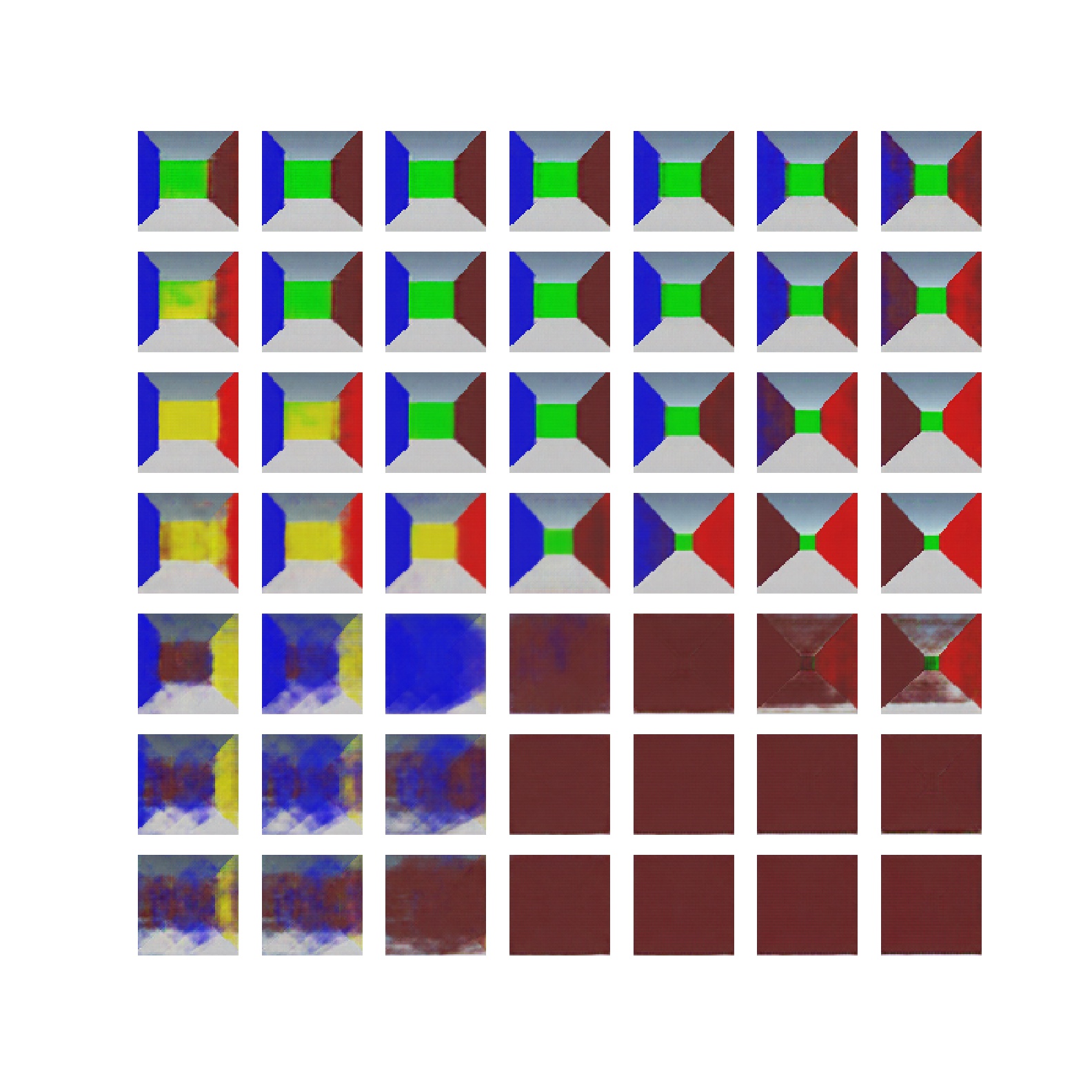}
  \end{minipage}

    \end{tabular}

  \caption{Example of the generated images from the latent vectors grid from the first and second principal components. Left: VAE ($\tau = 30$), Right: VAE/GAN ($\tau = 30$). }
  \label{fig:PCA_grid}
\end{figure}

To estimate the extent to which the mapped vectors are linearly projected, we compared the contribution of principal component analysis (PCA). We found that the cumulative contribution of PCA in VAE/GAN was higher than that of a single VAE (Fig.~\ref{fig:pca}) or projected into a lower dimensional space. The examples of the latent mapping $\{\bm{z}(t)\}$ in PCA space are shown in Fig.~\ref{fig:pca_mapping}, and the generated images from the latent vectors reconstructed from the grid that is reconstructed from the first two principal components are shown in Fig.~\ref{fig:PCA_grid}.
On careful inspection of the images generated from the PCA space, the image generated from VAE (left) tends to be blurred compared to that generated from VAE/GAN, and there seems to be less diversity in the generated images (Fig.~\ref{fig:PCA_grid}).

We further characterized the latent mapping $\{\bm{z}(t)\}$ in the PCA space by calculating two measures: smoothness $S_{\mbox{PCA}}$ and trajectory dissimilarity $d_{\mbox{LR}}$. Smoothness was evaluated as the average of the directional change and calculated as follows:

\begin{eqnarray}
S_{\mbox{PCA}} = \Big\langle\frac{ (\bm{z}_{\mbox{PCA}}(t)-\bm{z}_{\mbox{PCA}}(t-1))\cdot (\bm{z}_{\mbox{PCA}}(t+1)-\bm{z}_{\mbox{PCA}}(t)) }{\|(\bm{z}_{\mbox{PCA}}(t)-\bm{z}_{\mbox{PCA}}(t-1)  \| \|(\bm{z}_{\mbox{PCA}}(t+1)-\bm{z}_{\mbox{PCA}}(t)  \|}\Big\rangle,
\end{eqnarray}

where $\bm{z}_{\mbox{PCA}}(t)$ is the mapped latent vectors $\bm{z}(t)$ into the 2D PCA space.

\begin{figure}
  \centering
  \includegraphics[width=\linewidth]{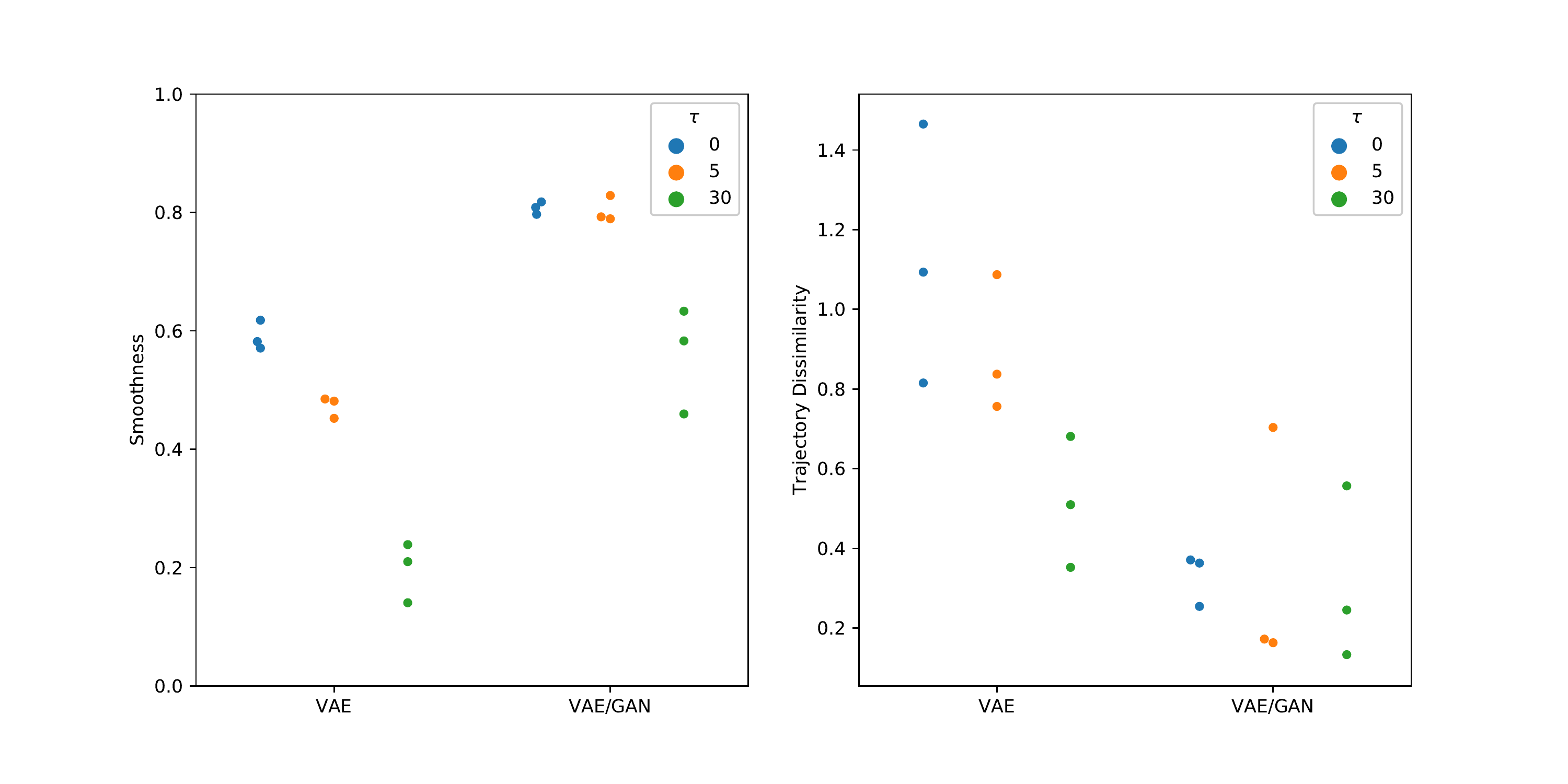}
  \caption{Characterization of the PCA maps. Left: Smoothness of the trajectory $\{\bm{z}(t)\}$ in the PCA space, $S_{\mbox{PCA}}$. Right: Trajectory dissimilarity between the right and left pathways of the 8-shaped environment, $d_{\mbox{LR}}$.}
  \label{fig:pca_features}
\end{figure}

Trajectory dissimilarity $d_{\mbox{LR}}$ is the median relative distance in the 2D PCA space between the latent trajectory of the left and right pathway of the environment. When we define the distance between $\bm{z}(t_1)$ and $\bm{z}(t_1)$ in the PCA space as,

\begin{eqnarray}
d_{\mbox{PCA}}(t_1,t_2) = \|(\bm{z}_{\mbox{PCA}}(t_1)-\bm{z}_{\mbox{PCA}}(t_2)\|,
\end{eqnarray}

then the trajectory dissimilarity $d_{\mbox{LR}}$ is calculated as,

\begin{eqnarray}
d_{\mbox{LR}} = \dfrac{\mbox{Median}(\{ d_{\mbox{PCA}}(t_{\mbox{L}}+\Delta t,t_{\mbox{R}}+\Delta t)\}_{0\leq\Delta t\leq T_{\mbox{path}}} )}
{\mbox{Median}(\{ d_{\mbox{PCA}}(t_{\mbox{L}}+\Delta t_1,t_{\mbox{R}}+\Delta t_2)\}_{0\leq\Delta t_1, \Delta t_2\leq T_{\mbox{path}}})},
\end{eqnarray}

where Median() is the median value, $t_{\mbox{L}}$ and $t_{\mbox{R}}$ are the initial times of the left and right paths of the 8-shaped environment, respectively, and $T_{\mbox{path}}$ is the duration of each pathway (Fig. ~\ref{fig:pca_features}).

We found that the smoothness of the trajectory was higher in VAE/GAN compared to that in VAE ($p < 0.001$, Tukey’s HSD). Also, the trajectories of the left and right pathways in the latent space were closer in VAE/GAN, especially when $\tau = 0$ ($p < 0.05$, Tukey’s HSD) (Fig.~\ref{fig:pca_features}). These results suggest that, with GAN, the movement direction in the latent space become stable and the two corresponding pathways, the left and the right, in the environment are similarly encoded in the latent space, which implies that the recognition of the environmental structure goes beyond the pixel similarity of the images, as shown in Fig.~\ref{fig:DistanceExample}. 

\begin{figure}
  \centering
  \includegraphics[width=\linewidth]{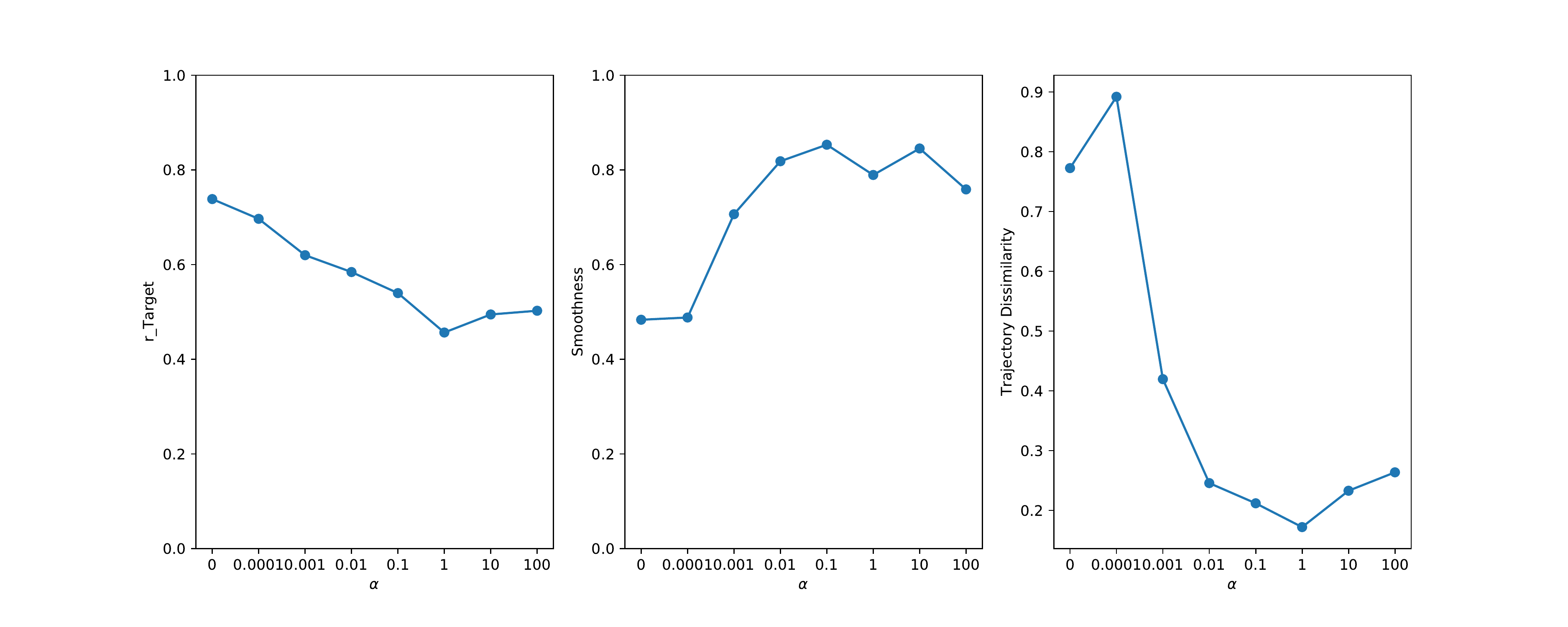}
  \caption{Dependence of the GAN weight parameter $\alpha$ on the resulting latent map feature. Left: Correlation coefficient between the distance matrix of the target images and latent vectors. Middle: Smoothness $S_{\mbox{PCA}}$ of the trajectory $\{\bm{z}(t)\}$. Left: Trajectory dissimilarity $d_{\mbox{LR}}$ between the right and left pathways of the 8-shaped environment.}
  \label{fig:param}
\end{figure}

To further investigate the contribution of GAN, we characterized the latent vectors resulting from the training of the loss function with different weighted $\alpha$ for the GAN loss function (Eq.~\ref{eq:loss_all_pixelLoss}). Here, we set $\tau = 5$. The results are shown in Fig.~\ref{fig:param}. First, we found that by increasing $\alpha$, the correlation between the distance matrix of the target images and the latent vectors, which corresponds the results in Fig.~\ref{fig:corr_distance}, decreased. This indicates that as the weights for GAN increases, the latent mapping does not simply reflect the pixel similarity of the images. Second, the smoothness of the trajectories $S_{\mbox{PCA}}$ was enhanced with the increase in GAN weighting. Therefore, by introducing GAN, the direction of the movement in the latent space stabilized, which suggests novel coordination (i.e., global coordinates), which is different from the VAE, which encodes each image separately. Third, we observed a decrease in the trajectory dissimilarity $d_{\mbox{LR}}$ with the increase in $\alpha$. Thus, by introducing GAN, the network recognizes the left and right pathways as alike, although not so similar in terms of the pixels of the corresponding images, which suggests a more abstract grasp of the environment. 
With too much weight on GAN loss, $d_{\mbox{LR}}$ started to increase, because, we speculate that, the mapping becomes inaccurate because there is less weighting on $\mathcal{L}_{\mbox{llike}}^{\mbox{pixel}}$ (Eq.~\ref{eq:loss_vae}).

In our dataset, the junction in the 8-shaped environment was different from the other part of the environment in such a way that the future outcome has two possibilities depending on whether the agent turns left or right. We investigated that how each network deals with these possible multiple outcomes (Fig.~\ref{fig:bif}). We found that each network has different coding schemes to deal with these multiple possibilities. In the case of VAE, the network ends up generating the superposition image of the two possibilities at the junction, which seems to minimize the pixel loss between the generated image and the images of two possibilities. $\mbox{VAE/GAN}$ also shows this tendency, but due to the presence of GAN, the generated image is not a simple superposition of the two possible outcomes, but it transformed in a way that it can deceive the Dis. In contrast, $\mbox{VAE/GAN}_{layerLoss}$, which corresponds to the original VAE/GAN architecture (Appendix.1), did not generate the superpositioned image, because the VAE/GAN evaluated the generated image based on the activation of the middle layer of the Dis, so the simple superposition did not minimize the loss function at the junction. Instead of the superposition, $\mbox{VAE/GAN}_{layerLoss}$ output the image of each possibility, but both possibilities are encoded in the neighbors of the other in the latent space (Fig.~\ref{fig:bif}).

\subsection{Sequence Generation by Closed Loop}

\begin{figure}
  \centering
  \includegraphics[width=\linewidth]{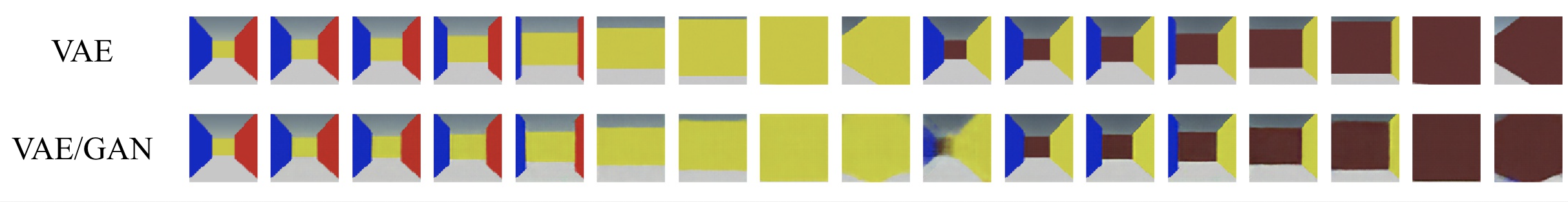}
  \caption{Examples of the generated image sequence by the closed loop from VAE and VAE/GAN ($\tau = 5$).}
  \label{fig:closedExample}
\end{figure}

After training the network to predict the upcoming visual inputs, we can make the network autonomous by using the following procedure. First, the initial image $\bm{x_0}$ is converted into a latent space vector by the Enc, $\bm{z_0} = \mathrm{Enc}(\bm{x_0})$, and an image is generated from the latent space vector by the Gen, $\bm{x_1} = \mathrm{Gen}(\bm{z_0})$. This image is now used as the input image. By repeating this process recursively, the network continues to predict the next image without receiving the actual image. We call this method a "closed loop" because it closes itself independent of any external input. We believe that this corresponds to, for example, the act of dreaming, or something like that. An example of the generated image sequence is shown in Fig.~\ref{fig:closedExample}.

A closed loop can be regarded as a dynamical system that converts an input image $\bm{x}$ into $\mathrm{Gen}(\mathrm{Enc}(\bm{x}))$ in a deterministic manner. Usually the Enc outputs the latent space vectors in a stochastic manner by sampling them from the Gaussian distribution with the parameters from the Enc output. However, here, we used the mean value of the Gaussian distribution as the output of $\mathrm{Enc}(\bm{x})$ to keep the closed loop procedure as deterministic. We followed this procedure from the input images for every 5 steps and iterated 200 times for each condition. The examples of the generated images from 180 to 200 steps are shown in Fig.~\ref{fig:longLoop}. We denote the image and the latent vector at the $i$ th iteration in the closed loop procedure as $\bm{x}_i$ and $\bm{z}_i$, respectively.

The resulting trajectories generated from the closed loop can be classified into three categories: fixed point, limit cycle, and chaotic dynamics. Trajectories converged to a fixed point were detected by calculating whether the accumulated change in the latent vector after 175th iterations $\displaystyle \sum_{i=175}^{199}\| \bm{z}_{i+1}-\bm{z}_{i}\|^2$ was below the threshold; here, we set the threshold as $z_{thr}=10^{-5}$. The limit cycle was detected based on whether the latent vector similar to the latent vector in the final iteration was included in the trajectory ($\displaystyle \min_i(\|\bm{z}_{200}-\bm{z}_{i}\|^2) < 10^{-8}$), and was not classified as the fixed point. We estimated the Lyapunov exponents of the rest of the trajectories by using the method by Rosenstein et al. \cite{rosenstein1993practical} and classified the trajectories with positive Lyapunov exponents as chaotic trajectories.

\begin{figure}
  \centering
  \includegraphics[width=\linewidth]{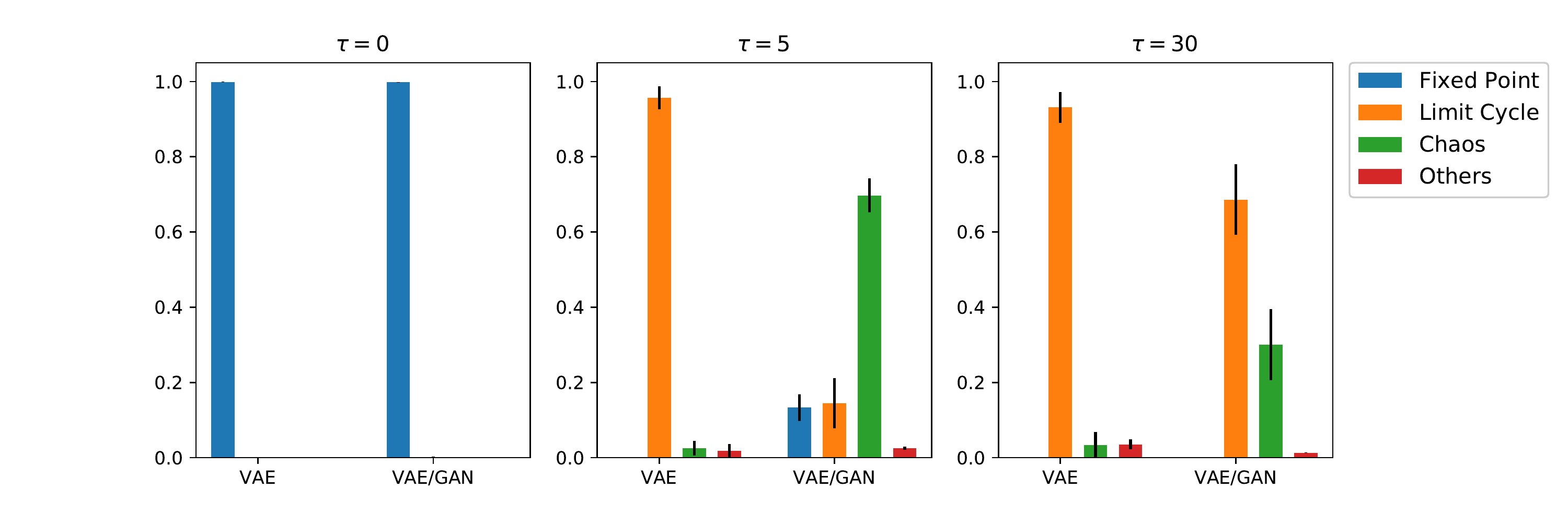}
  \caption{The ratio of each type of the closed-loop trajectories. The error bars indicate the standard deviation of the results from different random seeds ($n=3$).}
  \label{fig:TrajType}
\end{figure}

We classified the closed-loop trajectories for each conditions (Fig.~\ref{fig:TrajType}).
When $\tau = 0$, almost all trajectories converged to fixed points. When $\tau = 5, 30$, the trajectories from the VAE showed mainly limit cycle dynamics, while those from VAE/GAN showed an increased fraction of chaotic trajectories, which indicates that with GAN, the closed-loop dynamics was somewhat unstable and was able to generate novel sequences.

\section{Discussion}

In this paper, we presented an artificial cognitive map system based on a generative deep neural network and only using visual inputs. We aimed to show the alternative mechanism for the cognitive map that was different from path integration (i.e., the map was calculated from its own motion). Here, we argue that some aspects of the cognitive maps can be explained without the path integration. Our view is partly inspired by the recent findings of nonspatial coding in the hippocampal structures \cite{bellmund2018navigating}, which cannot be directly explained by the integration of the self-motion information.

First, we found that the distance of the dataset images (i.e., external world) was reflected in the distance structure of the latent vectors. The encoding of the proximity is one of the important ingredients for the cognitive map. Some studies, such as those focused on the encoding of social relationships \cite{tavares2015map} and those that use life-logging data \cite{nielson2015human}, focused only on the relationship between the distance structure in the data and the distance structure of the corresponding neural activity, so these mappings can be explained only by this aspect.
Another important finding was that when the network was trained to predict future images, the latent vectors reflected the distance structure of the predicted images, not the input images. This is in accordance with the fact that the cognitive map encodes not the present position but the predicted positions \cite{stachenfeld2017hippocampus}.

Second, by using GAN, we found that the trajectory in the latent space became smooth compared to that with only VAE.  This feature made the movement direction in the latent space stable, which means that when the agent moved in one direction, the corresponding latent vector also moved in certain direction. We regard this as the emergence of the global coordinates in the latent space. Also, we found that the trajectories corresponding to the left and right pathways were more alike in VAE/GAN. In the VR experiment conducted using rats \cite{aghajan2015impaired}, it was reported that when the rats always moved in a certain pathway, which they called the "systematic pillar condition," each segment of the pathway similarly activated the hippocampal neurons. This was interpreted as the encoding of the distance traversed, and this was enabled by the coupling of different modalities. Our result provides different perspectives to this. Each segment of the pathway can be mapped similarly if the agent can predict the upcoming visual input, and the latent representation was organized by GAN-like mechanisms. This phenomenon does not necessarily correspond to the "distance", which was more related to the path integration view, and does not require the coupling between different modalities.

Our system is based on a generative system, and we think that this is related to the other aspects of the hippocampus, such as the (episodic) memory. In the context of the episodic memory, Hassabis et al. \cite{hassabis2007using,hassabis2007patients} revealed that episodic memory and imagery shared the same neural basis. To be compatible with these two functionalities, the system has to be able to generate novel scenes while retaining the specific past memories. This resembles to our system considering that it can not only generate the past scenes but also produce novel and plausible scenes, which is enabled by GAN training. In the episodic memory study \cite{hassabis2007using}, it was reported that the activities of a specific brain region called precuneus was related to the familiarity of the visual experience, which might suggest that it functions like the Dis in GAN.

We also investigated the properties of the closed-loop trajectories. The closed-loop trajectories from VAE were stable and almost faithfully reproduce the experienced trajectories, but the trajectories from VAE/GAN were unstable, often showing chaotic dynamics, and we claimed that this might be the origin of the novelty in the "replay" in the hippocampus \cite{gupta2010hippocampal,stella2019hippocampal}. In this way, using a single model, we discussed the three aspects of the hippocampus: cognitive map, episodic memory and "replay". Usually, these are modeled independently using separate models, but we unified because they are actually implemented in the same region of the brain, hippocampus. The characteristics of the different functionality of the hippocampus should be correlated  with each other. For example, we speculate that the smoothness in the cognitive map is related to the novelty in the "replay". We hope that our study will provide the required information for the unified understanding of functionality of hippocampus.

\section{Conclusions}
We have constructed a novel artificial cognitive mapping system using generative deep neural networks, which can map input images to latent vectors and generate temporal sequences internally.
The results show that the distance of the predicted image after training is reflected in the distance of the corresponding latent vector. This indicates that the latent space is constructed to reflect the proximity structure of the data set, and may provide a mechanism through which many aspects of cognition are spatially represented \cite{behrens2018cognitive}. The present study allows the network to internally generate temporal sequences that are analogous to hippocampal replay/pre-play ability, where VAE produces only near-accurate replays of past experiences, but by introducing GANs, the latent vectors of the temporally close images are closely aligned and the sequences acquired some instability. This may be the origin of the generation of the new sequences that are found in the hippocampus
\cite{gupta2010hippocampal,stella2019hippocampal}.

\bibliographystyle{unsrt}  
\bibliography{references}  

\section*{Acknowledgement}

This work was partially supported by JSPS KAKENHI Grant, Correspondence and Fusion of AI and Brain (17H06024, 19H049790) and Chronogenesis : How the mind generates time (19H05306).

\newpage
\pagebreak
\appendix
\renewcommand{\thefigure}{A\arabic{figure}}
\setcounter{figure}{0}

\section{Appendix}

\subsection{VAE/GAN with Middle Layer Loss}

In the original VAE/GAN proposed in \cite{Larsen}, the reconstruction error of VAE/GAN was not the pixel loss between the input image and the generated image but was measured using the middle layer activation, which we call $\mbox{VAE/GAN}_{layerLoss}$. In this case, the loss function is given as follows:

\begin{equation}
  \mathcal{L}_{\mbox{VAE/GAN(layerLoss)}} = \mathcal{L}_{\mbox{prior}} + \mathcal{L}_{\mbox{llike}}^{\mbox{Dis}_l} + \mathcal{L}_{\mbox{GAN}},
    \label{eq:loss_all}
\end{equation}

with
\begin{eqnarray}
  \mathcal{L}_{\mbox{prior}} & = & \kld{q(\bm{z}|\bm{x})}{p(\bm{z})}\\
  \mathcal{L}_{\mbox{llike}}^{\mbox{Dis}_l} & = & -\expectation[q(\bm{z}|\bm{x}(t))]{\log p(\mbox{Dis}_l (\bm{x}(t+\tau))|\bm{z})}\\
  \mathcal{L}_{\mbox{GAN}} & = & \mbox{Dis}(\hat{x}) - \mbox{Dis}(\bm{x}) + \lambda \expectation[p(\bm{\hat{x}})]{(\| \nabla_{\bm{\hat{x}}}\mbox{Dis}(\bm{\hat{x}}) \|-1)^2},
    \label{eq:loss}
\end{eqnarray}

where $q(\bm{z}|\bm{x})$ is the Enc, $p(\bm{x}|\bm{z})$ is the Gen, $p(\bm{z}) = \mathcal{N}(\bm{0},\bm{I})$, $\bm{\hat{x}} \sim p(\bm{x}|\bm{z})$, and $p(\mbox{Dis}_l (\bm{x})|\bm{z})= \mathcal{N}(\mbox{Dis}_l (\bm{x})|\mbox{Dis}_l (\bm{\hat{x}}), \bm{I})$.

\subsection{Output Image Around the Bifurcation}

The example of the generated images at the bifurcation.

\begin{figure}[H]
  \centering
  \includegraphics[width=\linewidth]{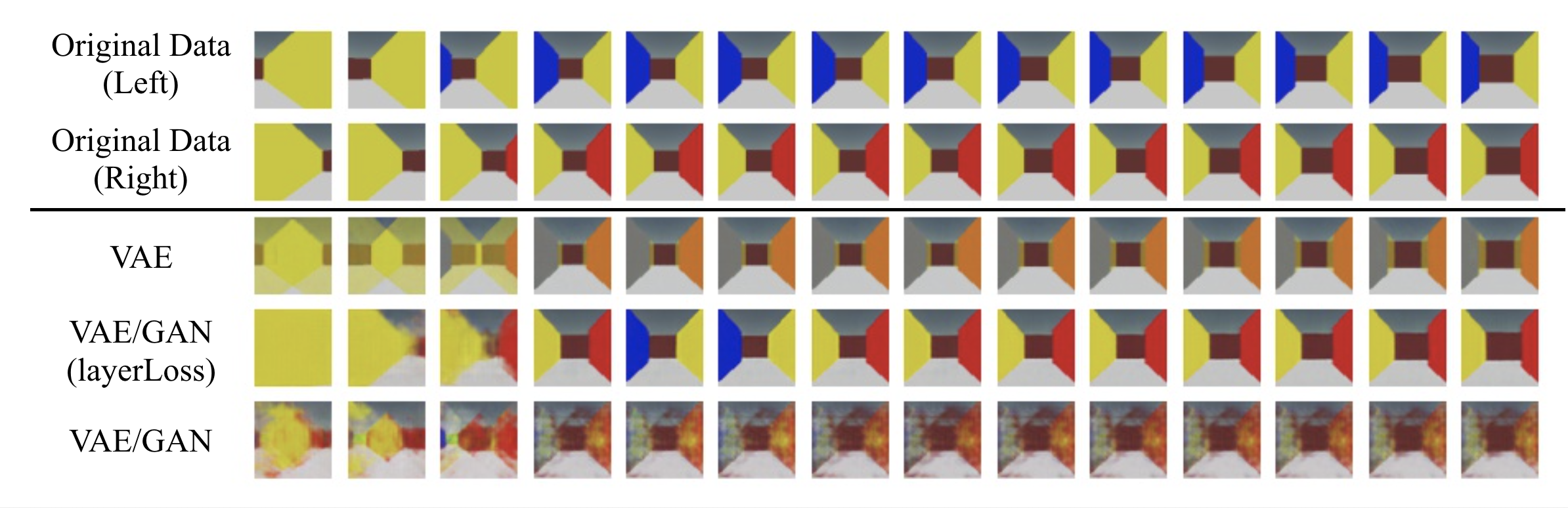}
  \caption{Examples of the output images around the bifurcation ($\tau = 30$). First and second row: The target images of left and right pathways, respectively. Third row: The generated images from VAE ($\tau = 30$).  Fourth row : The generated images from $\mbox{VAE/GAN}_{layerLoss}$ ($\tau = 30$). Fifth row : The generated images from VAE/GAN ($\tau = 30$). }
  \label{fig:bif}
\end{figure}

\subsection{Variability of the generated images from the same input image}

Using the encoder, we can stochastically sample the latent vectors and by putting these vectors the generator can output various outputs. Usually, this variation is relatively small and the generator outputs effectively the same image when provided the same input.

\begin{figure}[H]
  \centering
  \includegraphics[width=\linewidth]{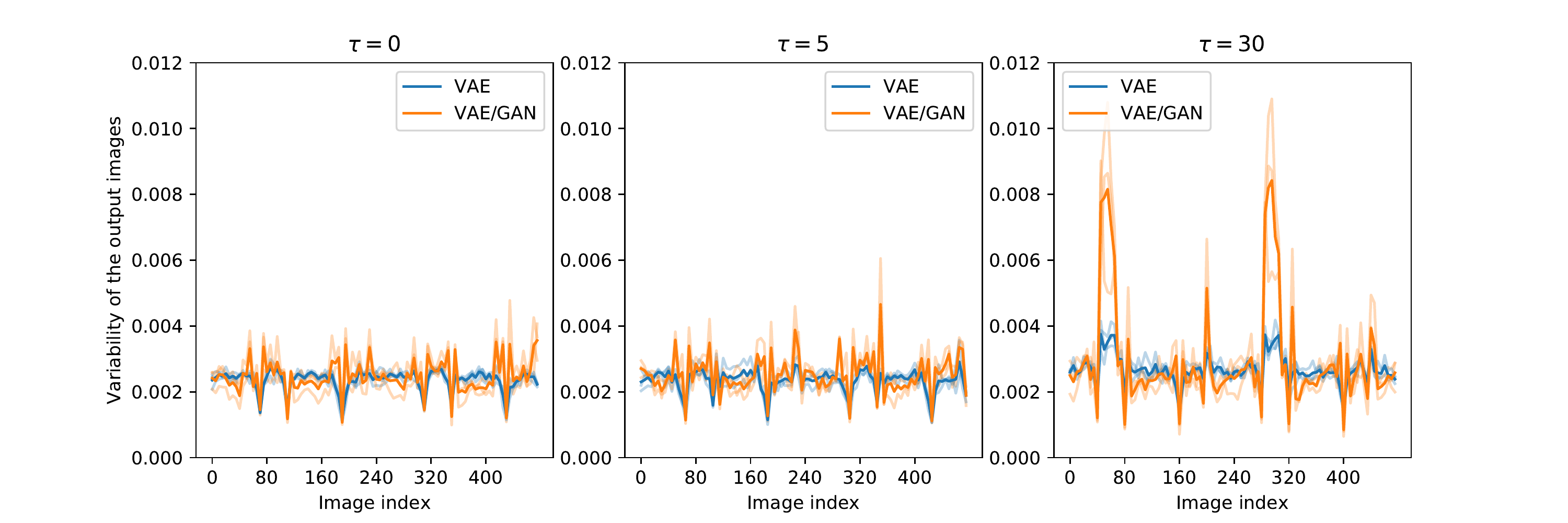}
  \caption{The standard deviation of the 10 images stochastically generated from the same input image. Image index 80 and 320 correspond to the bifurcation in the environment. The results from different random seeds were overlayed. }
  \label{fig:outImgStd}
\end{figure}

We investigated whether there were some input images which caused the large variation in the sampled output images (Fig.~\ref{fig:outImgStd}). We sampled the output images by sampling 10 latent vectors by providing the same input image and calculated the variability by taking the mean of the standard deviation of the pixel data. We found that when we feed the images before the bifurcation to the VAE/GAN with prediction timestep $\tau=30$, the variation of the output images were relatively large.

\subsection{Generated images from long step closed-loop generations}

The example of the generated images from the long step closed-loop generations.

\begin{figure}[H]
  \centering
  \includegraphics[width=\linewidth]{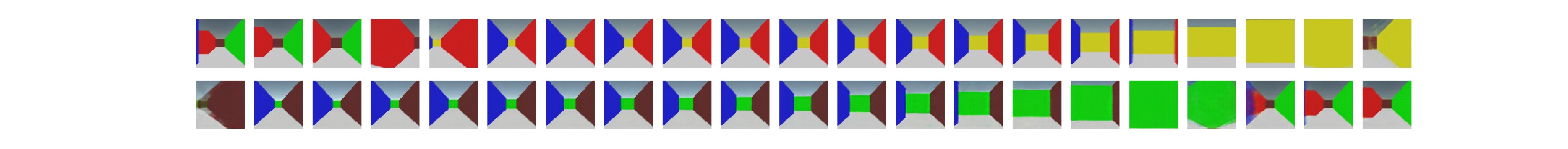}
  \caption{Examples of the generated image sequence after the 180 steps of the closed loop shown in Fig.\ref{fig:closedExample}. Top row: VAE, Bottom row: VAE/GAN.}
  \label{fig:longLoop}
\end{figure}

\end{document}